%% file: main_database.tex
\pdfoutput=1
\documentclass[10pt,a4paper,onecolumn]{article}
\usepackage[utf8]{inputenc}
\usepackage{amsmath}
\usepackage{amsfonts}
\usepackage{amssymb}
\usepackage{array}
\usepackage{fancyvrb}
\usepackage{afterpage}
\usepackage[section]{placeins}

\usepackage{lipsum}
\usepackage[T1]{fontenc}
\usepackage{multirow}
\usepackage{longtable,tabu}
\usepackage{caption}
\usepackage{graphicx}
\usepackage{xcolor}
\usepackage{rotating}
\usepackage{hyperref}
\usepackage{enumitem}
\usepackage{placeins}
\usepackage{subfig}
\usepackage[export]{adjustbox}

\usepackage[a4paper,bindingoffset=0.2in,%
            left=1.5cm,right=1.5cm,top=3cm,bottom=3cm,%
            footskip=.25in]{geometry}
            
\usepackage{bbm}
\usepackage{float}
\usepackage[bottom]{footmisc}
\usepackage{relsize}
\usepackage{titlesec}
\definecolor{blue}{HTML}{1A1C41}
\usepackage[longnamesfirst]{natbib}
\bibpunct{(}{)}{,}{a}{,}{,}
\usepackage{dirtytalk}
\usepackage{booktabs,threeparttable}

\usepackage{url}
\makeatletter
\newcommand{\thickhline}{%
    \noalign {\ifnum 0=`}\fi \hrule height 1pt
    \futurelet \reserved@a \@xhline
}
\newcolumntype{"}{@{\hskip\tabcolsep\vrule width 1pt\hskip\tabcolsep}}
\makeatother

\renewcommand{\thesection}{\Roman{section}} 
\renewcommand{\thesubsection}{\thesection.\Roman{subsection}}
\renewcommand{\thesubsubsection}{\thesubsection.\Roman{subsubsection}}

\titleformat{\section}
{\normalfont\large\bfseries}{\thesection}{1em}{}
\titleformat{\subsection}
{\normalfont\normalsize\bfseries}{\thesubsection}{1em}{}
\titleformat{\subsubsection}
{\normalfont\small\bfseries}{\thesubsubsection}{1em}{}

\usepackage{scalerel,stackengine}
\stackMath

\usepackage{multicol}
\usepackage{tgtermes}
\DeclareCaptionTextFormat{up}{\MakeTextUppercase{#1}}
\usepackage[font=small,labelfont=bf]{caption}
\usepackage{titling}
\begin{document}
\relscale{1}
\title{\textbf{HANA}: A \textbf{HA}ndwritten \textbf{NA}me Database for Offline Handwritten Text Recognition\thanks{Acknowledgements: We are grateful to the BYU Record Linking Lab for providing the US census data and the Copenhagen Archives who have supplied large amounts of scanned source material. The authors also gratefully acknowledge valuable comments from Philipp Ager, Anthony Wray, and Paul Sharp. \\The HANA database is available at \url{https://www.kaggle.com/sdusimonwittrock/hana-database}. Our code is available at \url{https://github.com/TorbenSDJohansen/HANA}.}}
\author{Christian M. Dahl$^1$, Torben Johansen$^1$, Emil N. Sørensen$^2$, Simon Wittrock$^1$}
\date{%
    $^1$University of Southern Denmark\\%
    $^2$University of Bristol\\[2ex]%
}
\maketitle
\begin{multicols}{2}
\input{./tex/0_abstract.tex}

\input{./tex/1_introduction.tex}

\input{./tex/2_database.tex}

\input{./tex/3_results.tex}

\input{./tex/4_discussion.tex}

\input{./tex/5_conclusion.tex}

\bibliography{bib}

\newpage
\input{tex/7_Appendix}

\end{multicols}

\end{document}

%% file: tex/0_abstract.tex
\section*{Abstract} 
\textit{Methods for linking individuals across historical data sets, typically in combination with AI based transcription models, are developing rapidly. Probably the single most important identifier for linking is personal names. However, personal names are prone to enumeration and transcription errors and although modern linking methods are designed to handle such challenges, these sources of errors are critical and should be minimized. For this purpose, improved transcription methods and large-scale databases are crucial components. This paper describes and provides documentation for HANA, a newly constructed large-scale database which consists of more than 3.3 million names. The database contain more than 105 thousand unique names with a total of more than 1.1 million images of personal names, which proves useful for transfer learning to other settings. We provide three examples hereof, obtaining significantly improved transcription accuracy on both Danish and US census data. In addition, we present benchmark results for deep learning models automatically transcribing the personal names from the scanned documents. Through making more challenging large-scale databases publicly available we hope to foster more sophisticated, accurate, and robust models for handwritten text recognition} 

%% file: tex/1_introduction.tex
\section{Introduction}
\label{Introduction}
As part of the global digitization of historical archives, the present and future challenges are to transcribe these efficiently and cost-effectively. We hope that the scale, quality, and structure of the HANA database can offer opportunities for researchers to test the robustness of their handwritten text recognition (HTR) methods and models on more challenging, large-scale, and highly unbalanced databases. The availability of large scale databases for training and testing HTR models is a core prerequisite for constructing high performance models. While several databases based on historical documents are available, only few have been made available for personal names. For linking, matching, or genealogy, the personal names of individuals is one of the most important pieces of information, and being able to read personal names across historical documents is of great importance for linking individuals across e.g. censuses:  See, for example, \citet{Abramitzkyetal2012}, \citet{Abramitzkyetal2013}, \citet{Abramitzkyetal2014}, \citet{Abramitzkyetal2016}, \citet{Abramitzkyetal2020a}, \citet{Abramitzkyetal2020b},   \citet{Feigenbaum2018}, \citet{Massey2017}, \citet{Baileyetal2020},  and  \citet{Priceetal2019}. Importantly, \citet{Abramitzkyetal2020a} and \citet{Baileyetal2020} both discuss the rather low matching rates when linking transcriptions of the same census together. This is partly due to low transcription accuracy of names. Furthermore, both papers are concerned about the lack of representativeness of the linked samples. These two observations strongly motivates the HANA database, i.e., collecting and sharing more data of higher quality, and to the work on improving transcription methods in order to reduce the potential biases in record linking.

\vspace{0.2cm}
In total, the HANA database consists of more than 1.1 million personal names written on single-line images with each personal name consisting of an average of three names. All original images are made electronically available by Copenhagen Archives and the processed database described is made freely available.
While most of the existing databases contain single isolated characters or isolated words, such as the names available in the Handwriting Recognition database at Kaggle, one of the important features of our database is the resemblance with other challenging historical documents, where source data often contain general image noise, different writing styles, and varying traits across the images.\footnote{The Handwriting Recognition database on handwritten names, containing names with somewhat clearly separated characters, is available at \url{https://www.kaggle.com/landlord/handwriting-recognition}. Similarly to the transfer learning exercises we present in detail in Section~\ref{Results}, Appendix~\ref{app: Appen} shows that our approach also works for this larger dataset, although the performance increase is smaller.}


\vspace{0.2cm}
The rest of the paper is organized as follows: In Section \ref{Constructing the HANA Database}, we describe the database and the data acquisition procedure in detail. Section \ref{Results} presents the benchmark results on the database using a ResNet-50 deep neural network in three different model settings. In addition, this section provides examples validating how the HANA database can be used for transfer learning on, e.g., Danish and US census data. In Section \ref{Discussion}, we discuss the database and the benchmark methods and results, in addition to some considerations for future research. Section \ref{Conclusion} concludes.

%% file: tex/2_database.tex
\section{Constructing the HANA Database}
\label{Constructing the HANA Database}
\begin{figure}[H]
  \caption{\\Example of a Police Register Sheet}
  \label{fig:PRS_example}
  \centering
  \captionsetup{justification=justified}
  \includegraphics[width=0.4 \textwidth]{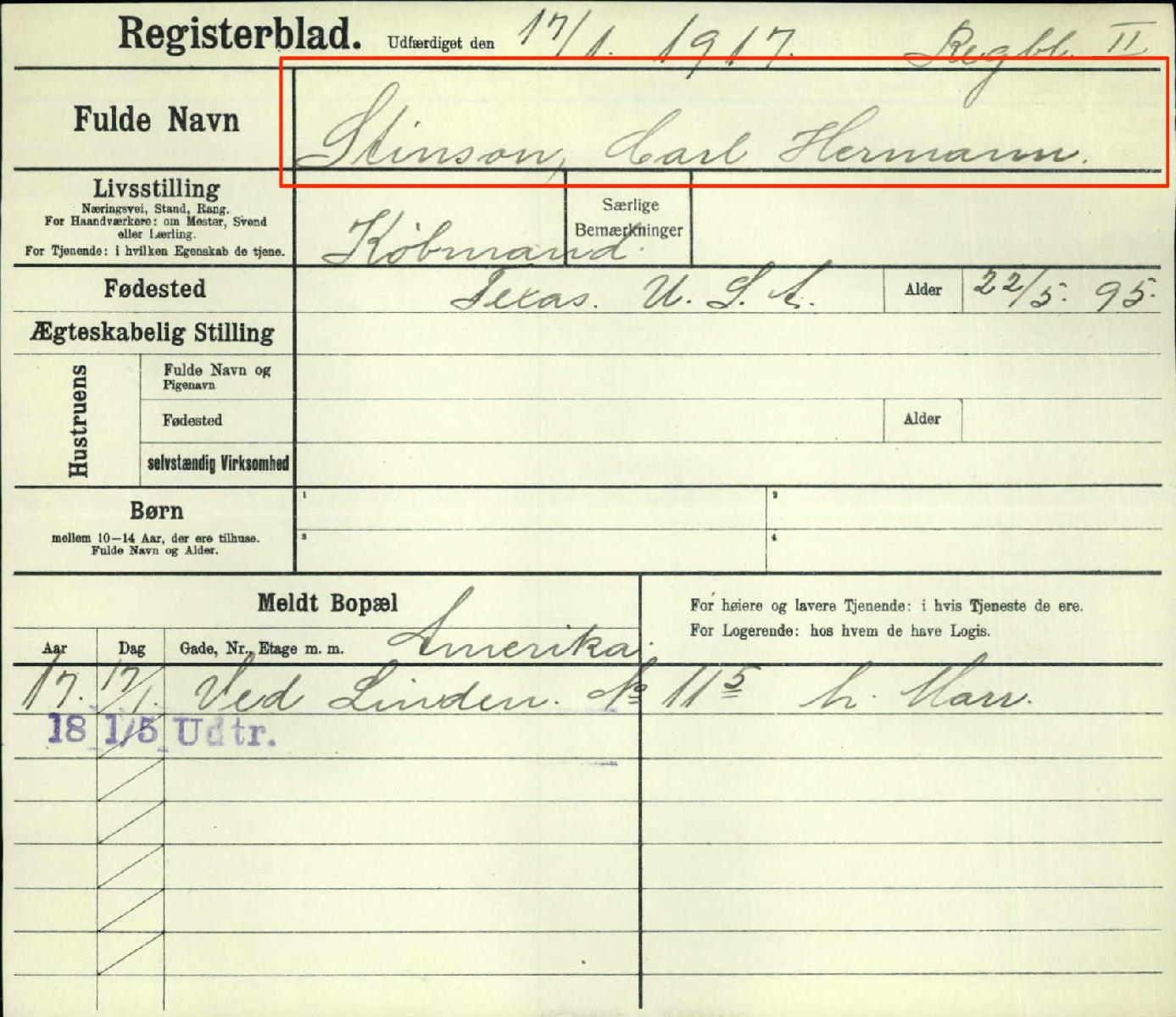}
  \caption*{\scriptsize{The figure shows an example of the raw documents received from Copenhagen Archives. The first line specifies the date the document was filled. The second line contains the full name (which we have outlined by adding a red bounding box not present in the original images) while the occupation is written in the smaller region just below the name. The fourth line contains the birth place and the birth date. The sections below contain information on the spouse and children.}}
\end{figure}

This section describes the HANA database in detail and the image-processing procedures involved in extracting the handwritten text from the forms. In 1890, Copenhagen introduced a precursor to the Danish National Register. This register was organized and structured by the police in Copenhagen and has been digitized and labelled by hundreds of volunteers at Copenhagen Archives. In Figure \ref{fig:PRS_example}, we present an example of one of these register sheets.

\paragraph{The Register Sheets} In total, we obtain 1,419,491 scanned police register sheets from Copenhagen Archives.  All adults above the age of 10 residing in Copenhagen in the period 1890 to 1923 are registered in these forms. Children between 10 and 13 were registered on their father's register sheet. Once they turned 14, they obtained their own sheet. Married women were recorded on their husband's register sheet, while single women were recorded on their own sheet. This is most likely biasing our database to include more men than female, as we focus on the main individual on the register sheets. The female to male ratio calculated based on the number of spouses registered is somewhere between 1 and 1.5 in the final database with the most likely male to female ratio being 1.3 (56\% men relative to 44\% women). 

\vspace{0.2cm}

Prior to 1890, the main registers used by the police were the census lists which goes back to 1865 and lasted until 1923. However, due to the census lists only being registered twice a year, in May and November, some migration across addresses was not recorded, and individuals residing only shortly in the city would not have been recorded \citep{stadsarkiv_mandtal}.



\vspace{0.2cm}

A wealth of information is recorded in the police register sheets, including birth date, occupation, address, name of spouse, and more, all of which is systematically structured across the forms. While this paper focuses on extracting and creating benchmark results for the personal names, the remaining information can be constructed using similar procedures to those presented in this paper and may serve as additional databases for HTR models.

Rare information is also included in the register sheets, such as whether the individuals were wanted, had committed prior criminal offences, or owed child benefits. This kind of information is written as notes and is therefore typically written under special remarks in the documents. As opposed to the censuses, which were sorted by streets and dates, the police register sheets were sorted by personal names. This made it easier for the police to control the migration of citizens of Copenhagen and track individuals over time. Once an individual died, they were transferred to the death register \citep{deathregister}.

\vspace{0.2cm}

In 1923, the Danish National Register replaced the registration of all citizens in Copenhagen \citep{stadsarkiv}. Ever since 1924, the Danish National Register has registered all individuals in all municipalities in Denmark \citep{folkeregister}.

\paragraph{Data Extraction and Segmentation} To segment the data, we use point set registration. Point set registration refers to the problem of aligning point spaces across a template image to an input image \citep{registration}. To find point spaces that roughly correspond to each other across semi-structured documents, we extract horizontal and vertical lines from the document. We use the intersections as the point space, which we align with the template points. We briefly outline the method below; see \citet{dahl2020} for more details.

\vspace{0.2cm}

To start the process of extracting the personal names from the forms, we binarize the images. We extract horizontal and vertical lines from the documents by performing several morphological transformations, see, e.g., \citet{extracting_lines}. The intersections are subsequently found using Harris corner detection \citep{harris}. Once we have the point space defined, we use Coherent Point Drift \citep{CPD}, which coherently aligns the point space from the input image to the point space on the template image. This yields a transformation function that maps the points found in the input images to the points in the template image. To improve the segmentation performance of the database, we add several restrictions to the transformations such that all extreme transformations are automatically discarded. This reduces the size of the database to just over 1.1 million images with attached labels. Even though this removes more than 20\% of the data, we believe the gain from more reliable data outshines the cost associated with a smaller database. 

\vspace{0.2cm}

Once we have prepared the images, we clean the labels to fit into a Danish context, which implies that all non-Danish variations of letters are replaced with the Danish equivalent of these. A few of these might be incorrect, e.g., if the individuals are foreigners, but we expect the level of mis-classification arising from this to be smaller than the number of characters labelled incorrectly by the volunteers at Copenhagen Archives. In addition, we restrict the sample to names that only contain alphabetic characters and with a length of at least two characters, yielding a final database of 1,105,904 full names.

\vspace{0.2cm}

It is possible to increase the number of extracted names for each sheet by considering the spouse and children of an individual. However, this would entail lowering the quality of the data, as the last name is not necessarily present for these individuals and the quality of the segmentation is also lower. Hence, we leave this for future work.
\vspace{0.2cm}

The personal name labels are either categorized as first or last names by Copenhagen Archives. Most commonly, the last name is written as the first word on the image while the subsequent words are the first and middle names (in that order). However, some exceptions occur, and there are other rules that may interfere with the structure of the ordering, such as underlining and numbering. The structure of the database can therefore be challenging for HTR models, as this structuring complication has to be overcome by the models.

\begin{figure}[H]
  \caption{\\Examples from the HANA Database}
  \label{fig:HANA_example}
\centering
\captionsetup{justification=justified}
  \includegraphics[width=0.4 \textwidth]{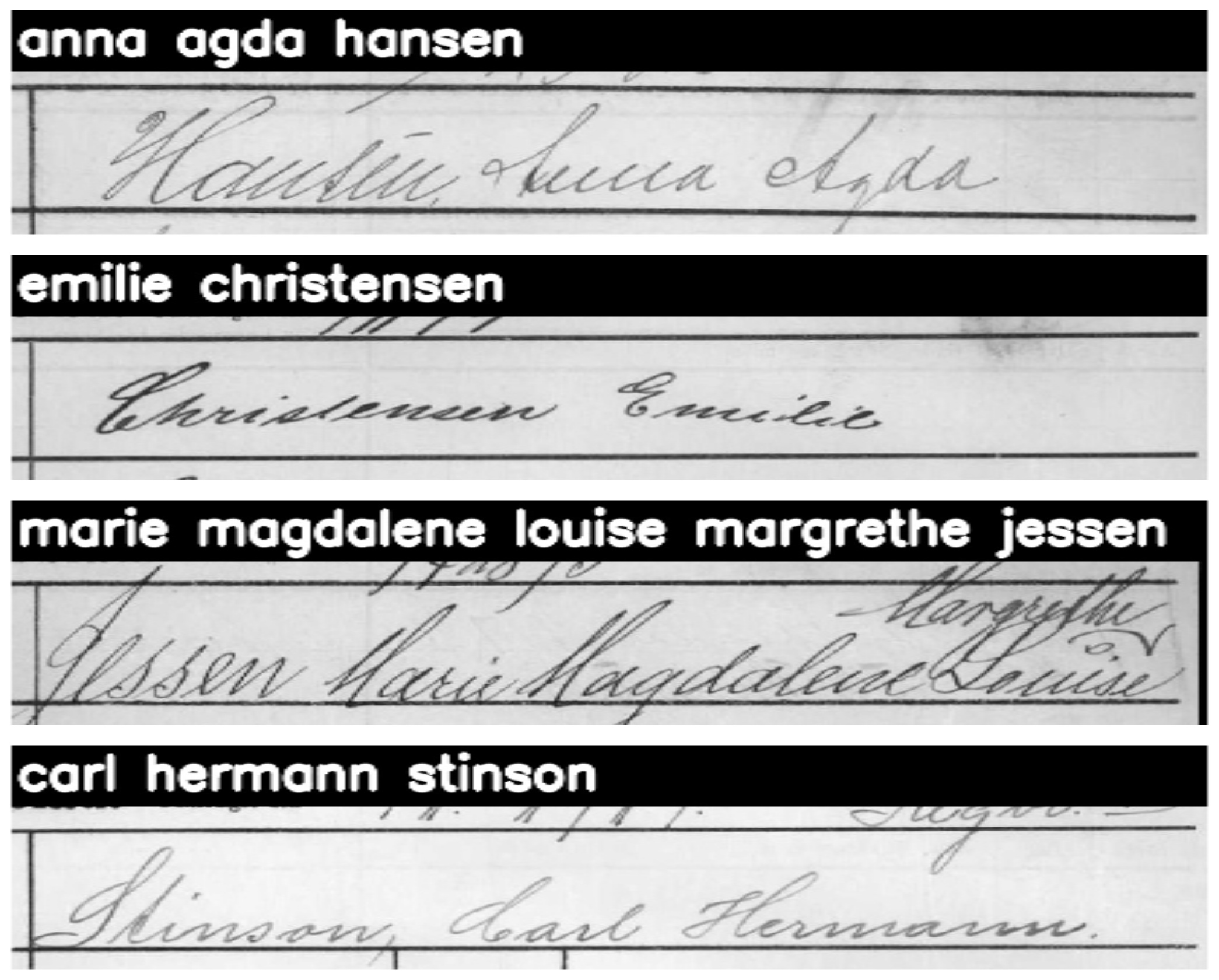}
\caption*{\scriptsize{The figure shows examples from the HANA database with the corresponding labels written above. The last name is typically written as the first word followed by the first and middle names, which is the case for all images above.}}
\end{figure}

\paragraph{Train and Test Splits} The test database consists of 5\% of the total database and is randomly selected. The training data consists of 1,050,082 documents while the test data consists of 55,822 documents. 2,129 surnames are only represented in the test sample, which contains a total of 10,228 unique last names relative to the overall of almost 70,000 unique last names.

\vspace{0.2cm}

As mentioned previously, the database is highly unbalanced due to vast differences in the commonness of names. Only the 604 most common surnames in the database occur at least 100 times, and only the 3,463 most common surnames occur more than 20 times. This covers slightly more than 85\% of the data, meaning that almost 15\% of the images contain names that occur fewer than 20 times. This naturally leads to challenges for any HTR model, as it needs to learn to recognize names with very few or even zero examples in the training data. However, this is also an important and indeed crucial goal to work towards.

\paragraph{Labelling} While transcribers at Copenhagen Archives were instructed to make accurate transcriptions of the register sheets, there exist humanly introduced inconsistencies in the labels. The same points made by \citet{deng2009imagenet} can be made here, as there are especially two issues to consider. First, humans make mistakes and not all users follow the instructions carefully. Second, users are not always in agreement with each other, especially for harder to read cases where the characters of an image are ``open to interpretation''.

\vspace{0.2cm}

With respect to the first point, we perceive this as part of the challenge for constructing any digital handwriting database, as they are all based on human transcriptions. For this database, Copenhagen Archives used super users to validate the transcriptions. In addition, it is possible to send requests for corrections at the website of Copenhagen Archives and thereby change incorrect labels. With respect to the second point, a number of considerations should be taken into account. A common labelling error found in the database is the existence of subtle confusing characters, similar names, or phonetically spelling errors. Characters or names that are often misread are, e.g., \textit{Pedersen} versus \textit{Petersen}, \textit{Christensen} versus \textit{Christiansen}, and \textit{Olesen} versus \textit{Olsen}. Solutions for these complications are difficult, as it is in many cases a judgement call by the transcriber. To reduce the number of incorrect labels in the training database, one could consider combining similar names, but we refrain from adopting that strategy.

\paragraph{Further Characteristics of the Database} Despite there being 69,906 unique surnames and 48,394 unique first and middle names, the total number of unique names amounts to only 105,607, as there is an overlap between the two sub-groups. There are fewer than 50 thousand examples of the characters \textit{q}, \textit{w}, \textit{x}, \textit{z}, \textit{å}, and \textit{æ}. For \textit{q} and \textit{å}, there are fewer than five thousand examples. The vast majority of names contain four to nine characters, with only 6.35\% of the names being shorter or longer. Quite frequently reported for Danish last names is the fraction of names ending with \textit{sen}. For this database, 710,117 surnames end with \textit{sen}, which corresponds to 64.21\% of all last names in the database. 
Appendix~\ref{Appendix} provides additional characteristics of the names in the database.

%% file: tex/3_results.tex
\section{Benchmark Results}
\label{Results}
This section describes the benchmark results published together with the HANA database and the value of transfer learning is illustrated. We use a variant of a ResNet-50 network for estimating the benchmark results. We transcribe the surnames in a character-by-character classification fashion. The predictions are subsequently matched to the closest existing name. One could also consider the surnames as an entity and classify each word in a holistic sense. We imagine that this could be problematic due to the unbalanced nature of this database and the training samples does not contain all unique names. We train three neural networks, one to predict the last name, one to predict the first and last name, and one to predict the entire name, i.e. first, middle, and last names.

\vspace{0.2cm}

We start by describing the architecture, optimization, and other details of the neural networks used in the paper. Scripts for the implementations are all in Python \citep{10.5555/1593511} using PyTorch \citep{paszke2019pytorch}.

\paragraph{Network Architecture}
Each neural network uses a ResNet-50 with bottleneck building blocks \citep{he2016deep} as its feature extractor; the weights of the PyTorch version of ResNet-50 pretrained on ImageNet \citep{deng2009imagenet} are used as the initial weights. The neural networks differ only insofar as their classification heads differ.
Here, a method similar to the one described in \citet{goodfellow2013multi} is used, with the exception that the sequence length is never estimated. The weights (and biases) of the classification heads are randomly initialized. For the last name network, 18 output layers are used (names are at most 18 letters long), each with 30 output nodes (letters a-å as well as a ``no character'' option).
For the first and last name network, 36 output layers are used (2 names of at most 18 letters), each with 30 output nodes.
For the full name network, 180 output layers are used (up to 10 names of at most 18 letters), each with 30 output nodes.

\paragraph{Optimization}
All neural networks are optimized using stochastic gradient descent with momentum of 0.9, weight decay of 0.0005, and Nesterov acceleration based on the formula in \citet{sutskever2013importance}. The batch size used is 256 and the learning rate is 0.05.
The networks are trained for 100 epochs and the learning rate is divided by ten every 30 epochs.
The loss of each classification head is the negative log likelihood loss of the head, and the total loss is the average of the negative log likelihood loss of each head.

\begin{center}
    \begin{threeparttable}[!ht]
\begin{center}
\scriptsize
\caption{\\WACC on the HANA database}
\label{tab:CHAR}
\centering
\begin{tabular}{lccc}
\thickhline
 \textbf{Names} & \textbf{Data coverage} & \textbf{WACC} & \textbf{WACC with Matching} \\
\thickhline
 \textbf{Last Name} & 100\% & 94.33\% & 95.68\% \\
 \textbf{First and Last Name} & 100\% & 93.52\% & 94.79\%\\
 \textbf{Full Name} & 100\% & 67.44\% & 68.81\% \\ 
 \textbf{Last Name} & 90\% & 98.36\% & 98.41\% \\
 \textbf{First and Last Name} & 90\% & 97.29\% & 97.46\%\\
 \textbf{Full Name} & 90\% & 72.78\% & 74.10 \% \\ 
\thickhline
\thickhline
\end{tabular}
\begin{tablenotes}[flushleft]
      \item{\scriptsize{The table shows the test performance of the HTR models as measured by word accuracy (WACC). The data coverage is defined as the fraction of the test database the model is tested on (keeping predictions where the network is most confident). For the models with 90\% data coverage, we remove the 10\% of the test sample where the model is most uncertain. All models are trained on the full train database allowing the networks to learn primitives and characters from uncommon names.
      }}
    \end{tablenotes}
    \end{center}
\end{threeparttable}
\end{center}

\paragraph{Image Preprocessing}
Images are resized to half width and height for computational reasons (resulting in images of width 522 and height 80). The images are normalized using the ImageNet means and standard deviations (to normalize similarly to the pretrained ResNet-50 feature extractor).
During training, image augmentation in the form of RandAugment with $N=3$ and $M=5$ is used \citep{cubuk2020randaugment}; the implementation is based on \citet{RandAugmentGitHub}.

\paragraph{Prediction of Networks}
Some post processing of predictions is performed. Each layer is mapped to its corresponding character (the 29 letters and the ``no character'' option).
Then, for each name (i.e. sequence of 18 output layers), the ``no character'' predictions are removed and the remaining letters form the prediction.
Letting, $\theta_i$ denote the ``no character'' option for character $i$, this means that both [h, a, n, $\theta_4$, s, $\theta_6, ..., \theta_{18}$] and [h, a, n, s, $\theta_5, ..., \theta_{18}$] will be transformed to \textit{hans}.

\paragraph{Matching}
As an additional step, we also test the performance if we refine the predictions of the networks by using matching. In some cases, a list of possible names (i.e. a lexicon of valid outcomes) may be present, in which case this can be used to match predictions that are not valid to the nearest valid name.\footnote{When using matching, we define our lexicons by assuming that the sample of names available in our data represents all possible names, and thus our lexicons might be considered too perfect, which could lead to an upward bias of the performance we report when using matching.} Specifically, we use the procedure in the \texttt{difflib} Python module to perform this matching.

For the last name network, the predictions that do not fall within the list of valid last names are matched to the nearest last name.
For the first and last name network, a similar procedure is used separately for the first name and the last name.
For the full name network, a similar procedure is used separately for the first name, the up to eight middle names, and the last name.

\paragraph{Performance Measures}
To measure the performance of our networks, we focus on the word accuracy (WACC) of our models.
Thus, a name prediction is considered to be incorrect if a single character or more of the name is transcribed incorrectly, and thus the character error rates are significantly lower than the word error rates implicitly reported.\footnote{WACCs are reported and the word error rates are given as 1 minus WACCs.}
We consider performance both with and without using matching to a lexicon as a post processing step.
Further, we report performance at different levels of \textit{data coverage}.
As our networks report a measure of their confidence for each prediction, we can rank all predictions by this measure.
Then, we can calculate the WACC at, e.g., 90\% data coverage by removing the 10\% of predictions where the network is the least certain.
We believe this metric is interesting, as it might be used to, e.g., (1) select the predictions where a sufficient WACC is reached or (2) let humans assist in transcribing images that the network is particularly uncertain about.

\paragraph{Results}
Table \ref{tab:CHAR} show an overview of the performance. The data coverage is either at 100\% or 90\% and presents the word accuracy if we use all transcriptions in the test set compared to only testing on the 90\% of the test data on which the model is the most certain. There is a trade-off between data coverage relative to accuracy, which is the motivation for also showing the results using a threshold at the 90th quantile. The table represent three different models for character-by-character recognition. The first model predicts only the characters in the last name, the second model predicts the first name and the last name, and the third model predicts the full name sequence. All of them are trained on the full database. For the full name model, the number of names present in a person's predicted name is equal to the number of names in the corresponding label in 96.85\% of the cases.
Using the Levenshtein distance to calculate the character error rate of the predictions (without matching) we find error rates of 1.48\% for the last name network, 1.66\% for the first and last name network, and 11.82\% for the full name network. The word accuracy for the last name model is 94.33\% without matching; this drops to 93.52\% for the first and last name model. The full name model is evaluated on the full names of the individuals and has to take into account the correct ordering of the all names in order to obtain each name correct, which we believe explain the performance deterioration for our rather simple network, where the word accuracy drops to 67.44\%.

\begin{figure}[H]
  \caption{\\Performance on the HANA database: Last Name}
  \label{fig:performance0}
\centering
\captionsetup{justification=justified}
  \includegraphics[width=0.5\textwidth]{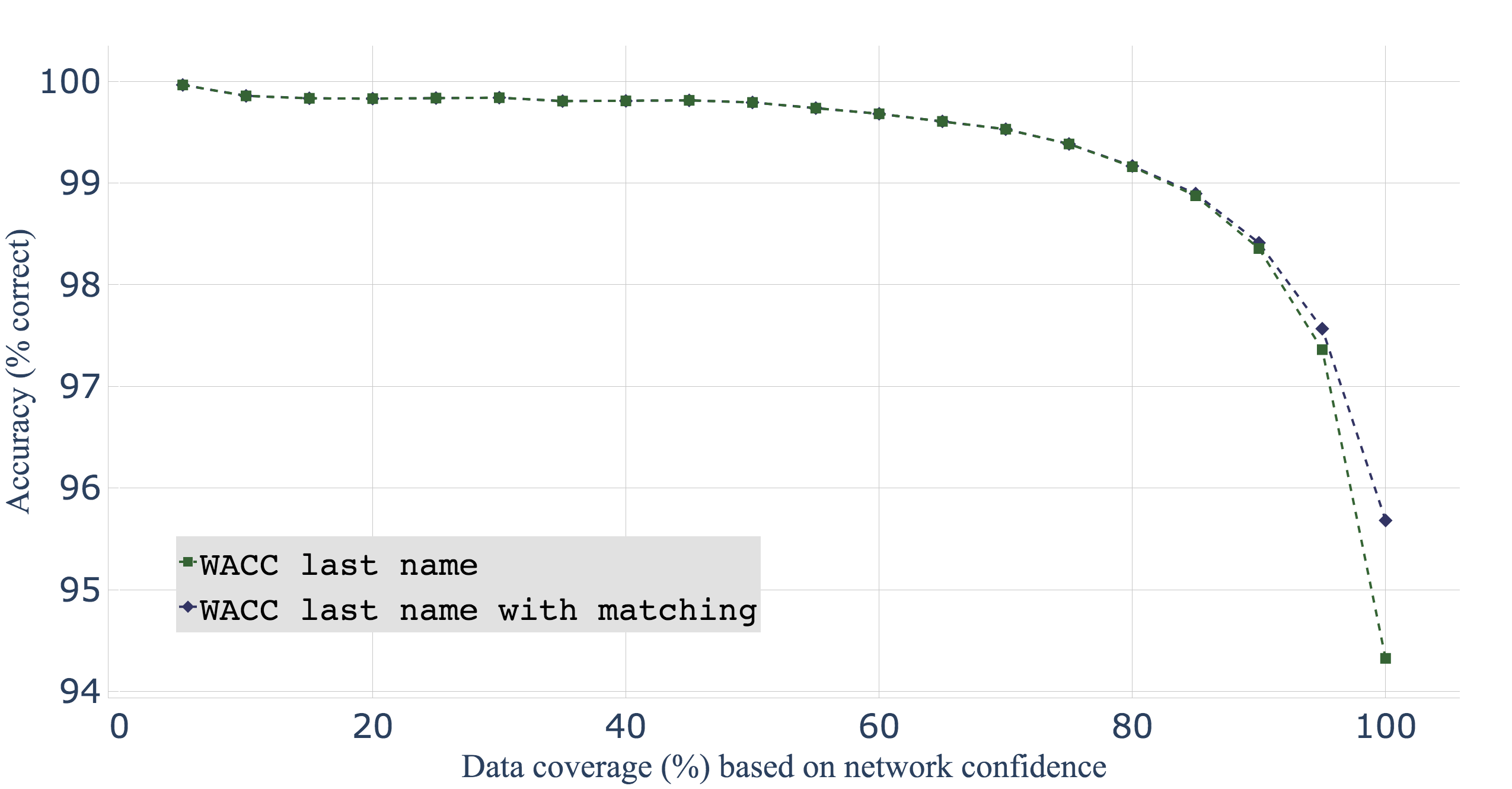}
  \caption*{\scriptsize{The figure shows the performance on the test set from the HANA database for the model trained on last names.
  We find that the matching of names to closest name relative to the unmatched performance is very similar until the 80th percentile. From this point on, the two lines diverge and the matched predictions clearly outperform the unmatched predictions.
  }}
\end{figure}

\begin{figure}[H]
\captionsetup{justification=centering}
  \caption{\\Performance on the HANA database: First and Last Name}
  \label{fig:performance1}
\centering
\captionsetup{justification=justified}
  \includegraphics[width=0.5 \textwidth]{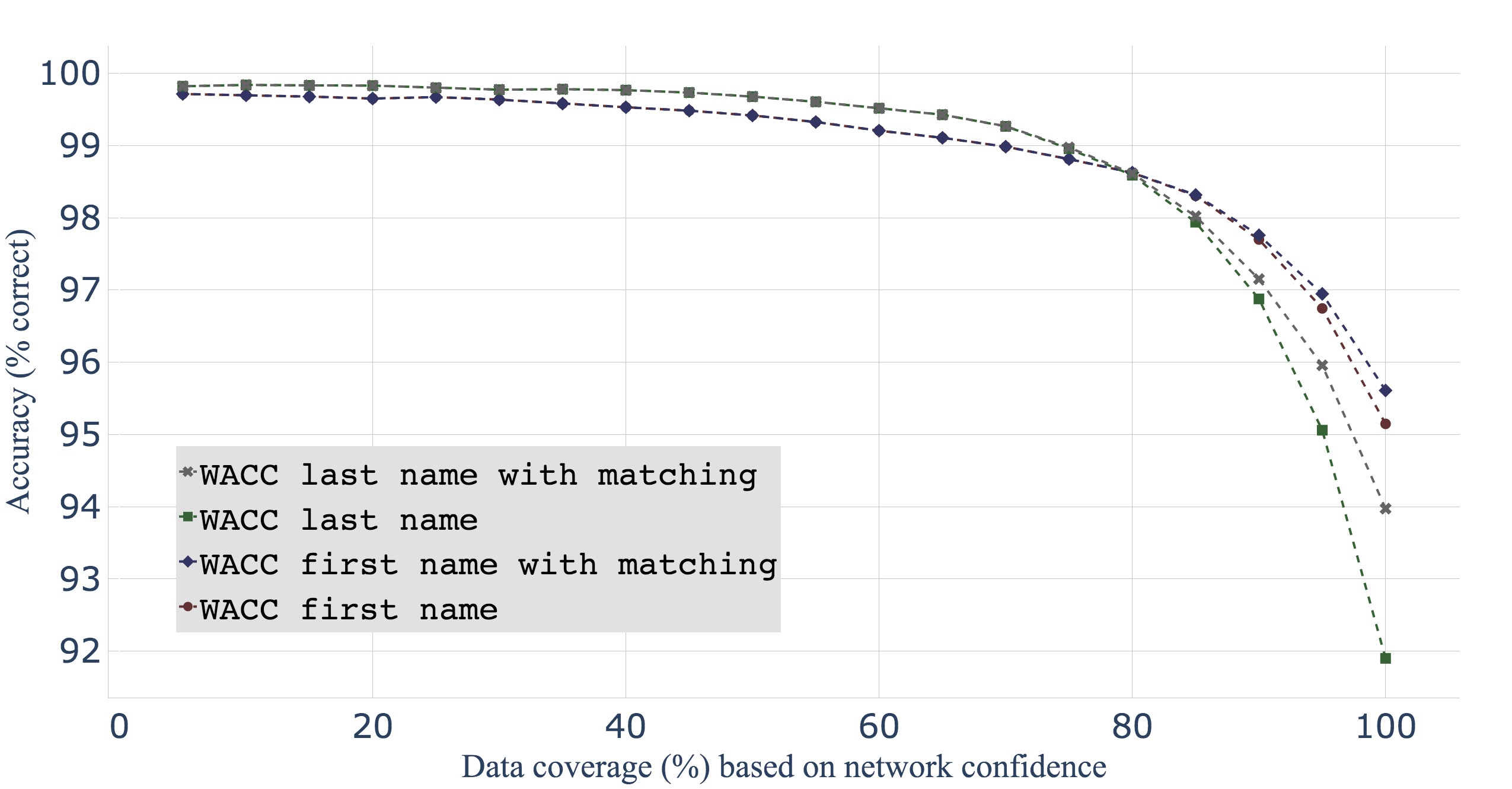}
  \caption*{\scriptsize{The figure shows the performance on the test set from the HANA database for the model trained on first and last names. We find that the network obtains a higher accuracy on the first names relative to the last names (reversing once below around 80\% data coverage) and that the last name accuracy is lower than the performance of the model that is trained only on last names.
  }}
\end{figure}

For the last name and the first and last name networks, Figures~\ref{fig:performance0} and \ref{fig:performance1} provide additional measures of the performances.
Figure~\ref{fig:performance0} shows the performance of the last name model over the entire range of data coverage, both with and without the use of matching. 
While both models achieve higher accuracy at lower data coverage, what is particularly interesting is the difference between the two: At full data coverage, matching improves the performance substantially, while this difference quickly disappears the moment the most uncertain predictions are sorted away.
Figure~\ref{fig:performance1} shows the corresponding performance for the first and last name network.
Here, the first and last name performances are illustrated separately. 
As for the last name model, it is evident that matching is particularly helpful at full data coverage and for last names.
Interestingly, while the network is better at transcribing first names at full data coverage, at around 80\% data coverage this changes, and it becomes better at transcribing last names, though the last name transcription performance of this model is lower than for the last name model, suggesting that estimating a separate model only for first names and using it in conjunction with the last name model might be superior to a model estimating both jointly.

\begin{figure*}
\centering
\caption{\\Examples from the Danish and US Censuses}
\label{fig:census_examples}
\subfloat[Danish Census]{
  \label{fig:census_examples-a}
  \includegraphics[height=0.32
  \textwidth]{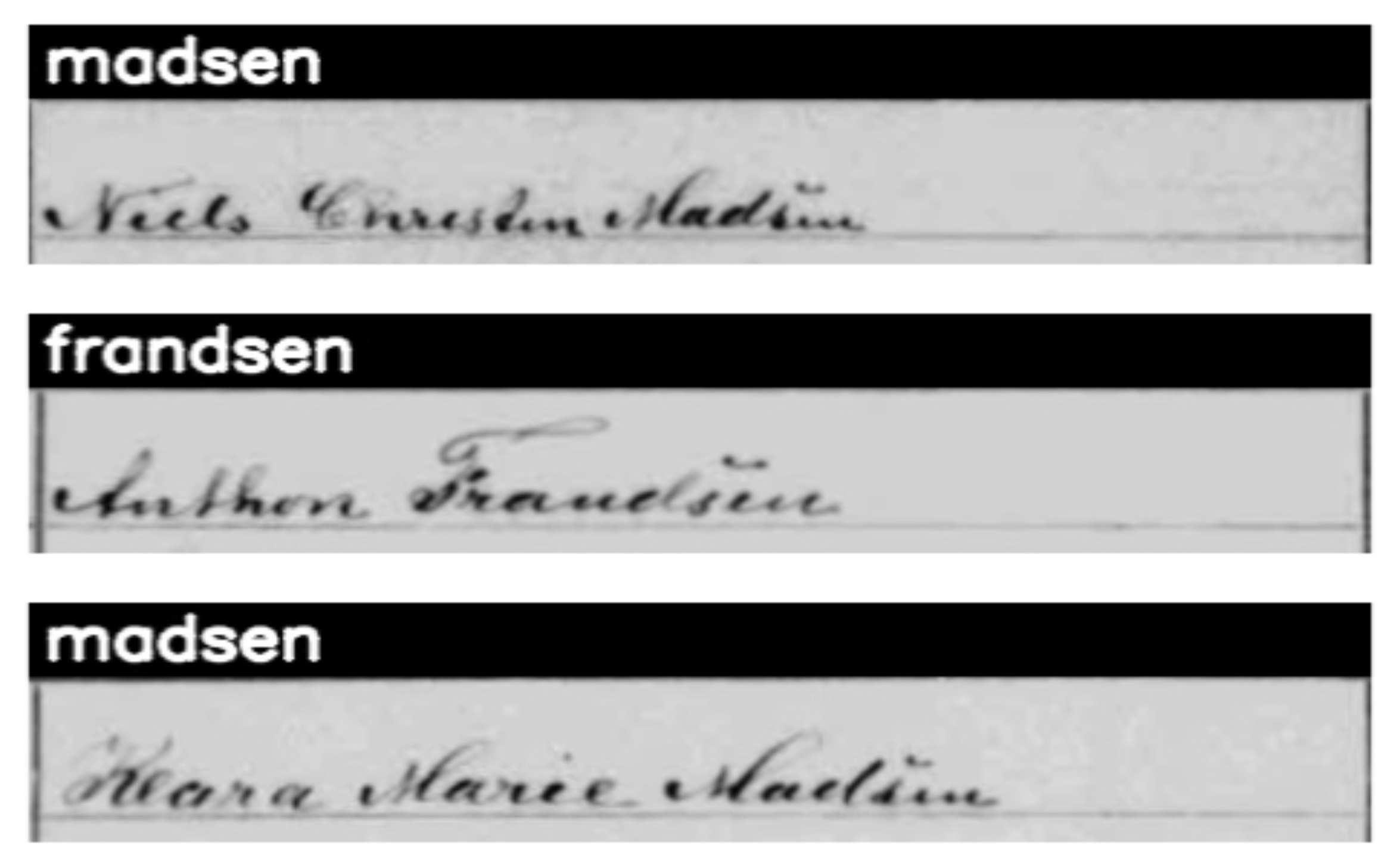}
}
\subfloat[US Census]{
  \label{fig:census_examples-b}
  \includegraphics[height=0.32\textwidth]{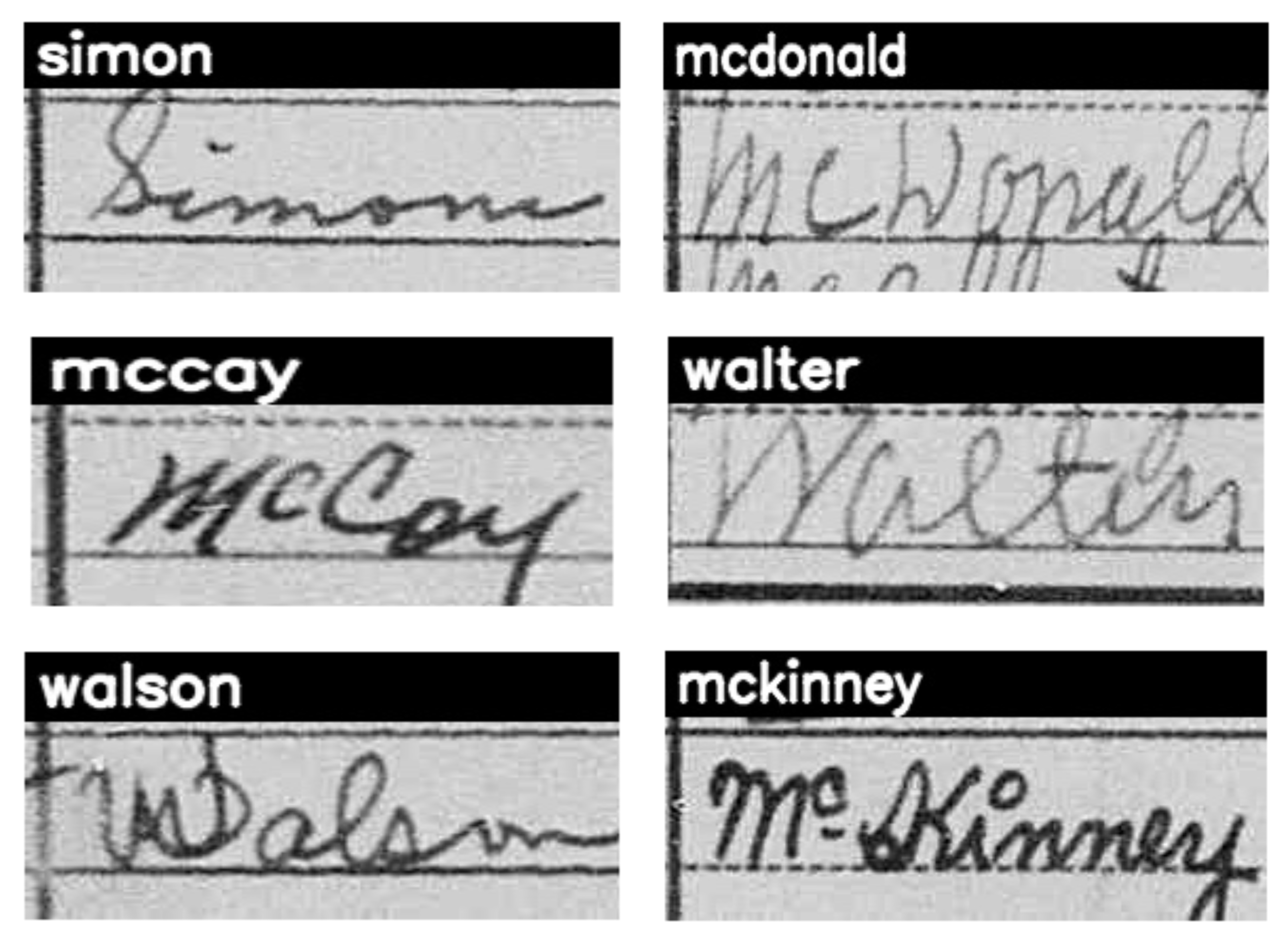}
}
\captionsetup{justification=justified}
\vspace{-0.2cm}
\caption*{\scriptsize{The figure shows examples of the surnames from the (Panel~\ref{fig:census_examples-a}) Danish and (Panel~\ref{fig:census_examples-b}) US census with the last names included in the text box above each minipic. 
As seen from these examples, the Danish census images mimics to a greater extent the images in Figure~\ref{fig:HANA_example} while the US census minipics include only the surnames of the individuals. However, it seems that in the Danish census the surname tend to be located to the right on the images while in the HANA database the surnames are usually located to the left, see Figure \ref{fig:HANA_example}. The width to height ratio of the HANA database is 6.5 which is similar to the Danish census with a ratio of 7.2 while the US census ratio is 3.7.}}
\end{figure*}

\begin{figure*}
\centering
\caption{\\Transfer Learning Performance on Danish and US Census}
\label{fig:transfer}
\subfloat[Danish Census]{
  \label{fig:transfer-a}
  \includegraphics[height=0.26
  \textwidth]{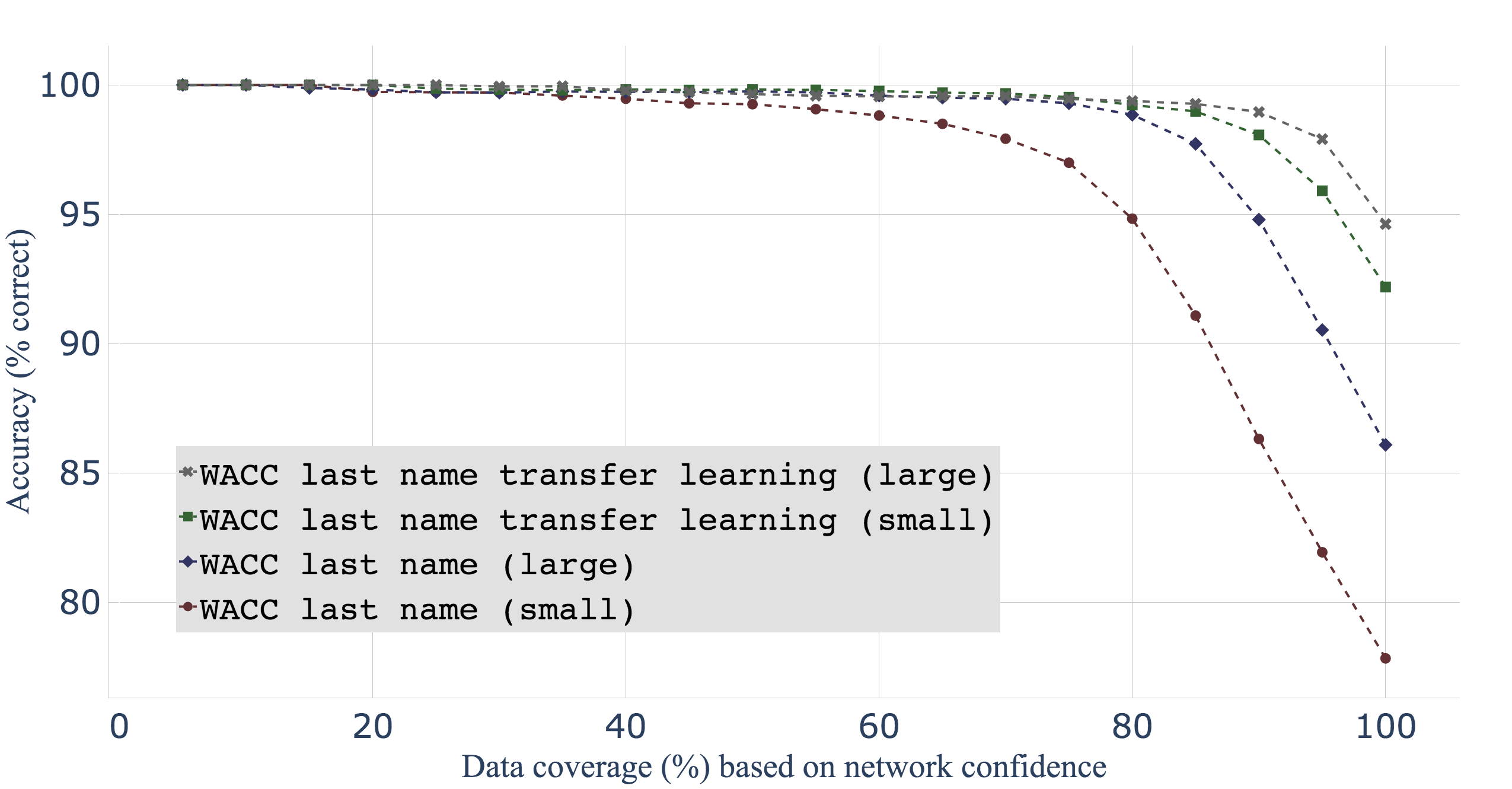}
}
\subfloat[US Census]{
  \label{fig:transfer-b}
  \includegraphics[height=0.26\textwidth]{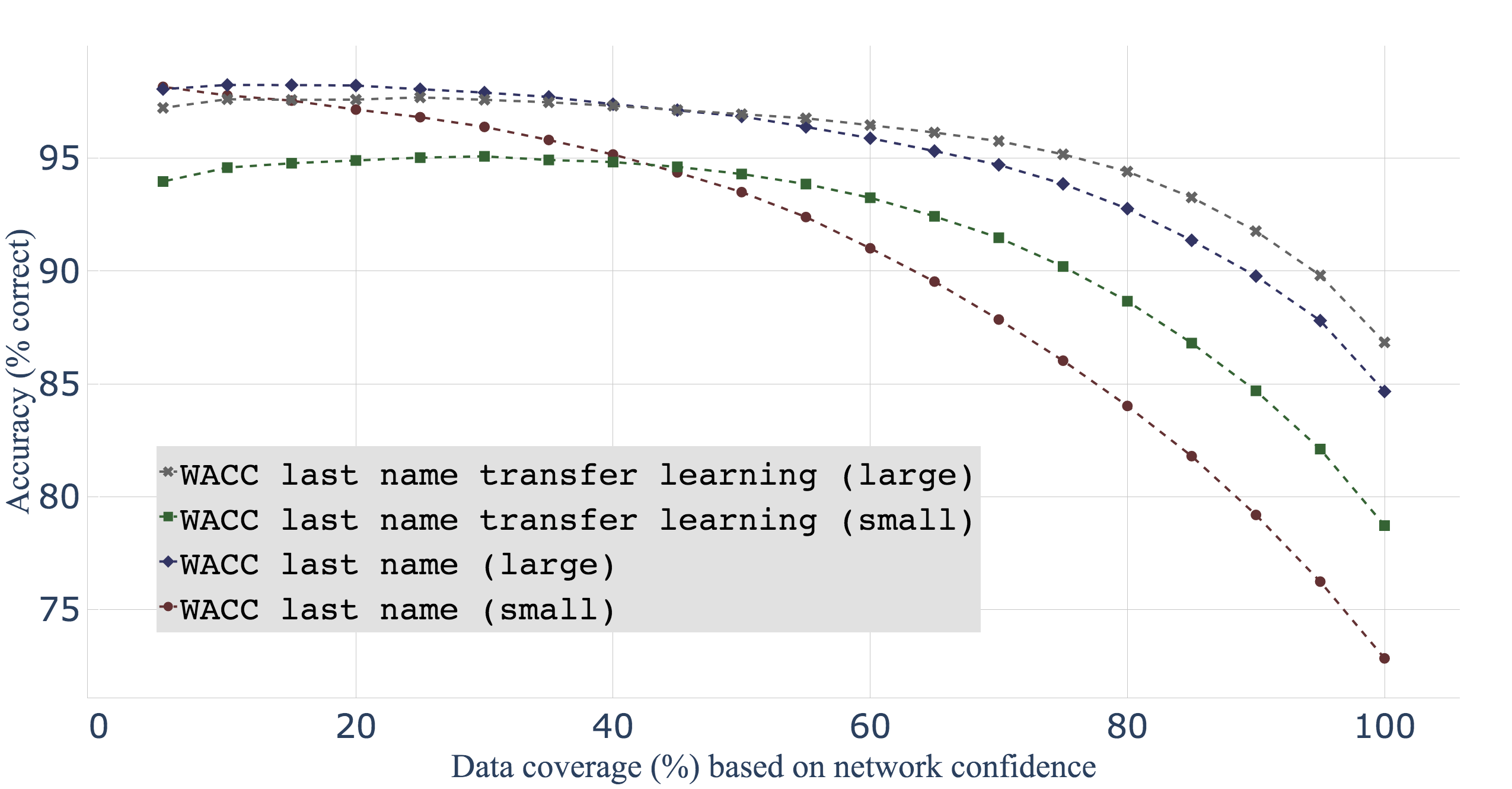}
}
\captionsetup{justification=justified}
\vspace{-0.2cm}
\caption*{\scriptsize{The figure shows the performance gain from adopting a transfer learning strategy based on the HANA database. 
Panel~\ref{fig:transfer-a} shows the performance on the Danish census data and Panel~\ref{fig:transfer-b} on the US census data.
We find that the performance gain is larger for the smaller training sets but still substantial with more than 50,000 training examples. Also, we find that the performance increase is larger for names that to a greater extent mimics the original HANA training examples, which is part of the reason for the better performance on the Danish census data. Most likely, the handwriting is also more similar across these datasets. Even so, the generability of the HANA database seems to be largely validated by the performance increase in both tested transfer learning exercises on the US census data.
  }}
\end{figure*}

\paragraph{Transfer Learning}
By publishing the database we aim to establish a foundation for transfer learning to handwritten names from other data sources.
This in turn can help others transcribe handwritten names more accurately -- while also reducing costs, as less manual labelling will be needed.
To motivate the usefulness of the HANA database for transfer learning, we present results for three separate transfer learning examples:
Transcription of handwritten surnames from Danish and US census data (see Figure~\ref{fig:census_examples} for some examples of these images), and transcription of the handwritten names from the Handwriting Recognition database from Kaggle which contain transcriptions of 410,000 handwritten names.
In this section, we provide details from our experiments for the two census datasets; Appendix~\ref{app: Appen} provides details for our third example.
In all cases, our results demonstrate that adopting a transfer learning strategy based on the HANA database can increase transcription accuracy, even when large amounts of training data is available.

\vspace{0.2cm}

For both the Danish and US census data, we present two sub-cases: We analyze the performance when a relatively small number of training images are available (approximately 10,000) and when a larger number of training images are available (approximately 50,000).
By training networks both with and without transfer learning on these datasets, we can infer the magnitude of the performance boost achieved by using the HANA database for transfer learning.

\vspace{0.2cm}

In total, we train eight new networks.
Due to the difference in the number of labelled images for each dataset and the use of transfer learning from the HANA database last name network, we expect that the optimal learning rate for each network might differ substantially.
For this reason, we perform a grid search on a validation set consisting of five percent of the training data for each network to tune the learning rate.
All other training settings are similar to those we used to train models on the HANA database.
Thus, all new models are similar to the last name model on the HANA database, and training only differ with respect to the learning rate used and the starting weights.\footnote{The image sizes of the source files also differ, and are around 465 by 65 for the Danish census and 350 by 95 for the US census, which are the resolutions we train these networks at.
To provide the most fair comparison, we use weights from pretrained models on ImageNet as the initial weights for the models where we do not transfer learn from the HANA database, similarly to our training of the models on the HANA database.}

\vspace{0.2cm}

Figure~\ref{fig:transfer} shows our main findings.
The performance based on the Danish census is illustrated in Panel~\ref{fig:transfer-a}, while the performance based on the US census is illustrated in Panel~\ref{fig:transfer-b}. 
The data coverage is gradually increasing along the first axis, and at a data coverage of 100\% it is clear that the worst performing model for each census is the network trained on the small database without transfer learning while the best performing model is the network trained on the large database with transfer learning.
Quite interestingly, there is a difference between which model is the second best between the Danish and US census.
For the Danish census, the model trained on the small database with transfer learning is better than the model trained on the large database without transfer learning, while this is reversed for the US census.
This is likely due to the larger similarity between the HANA database and the Danish census compared to the similarity between the HANA database and the US census (see Figures \ref{fig:census_examples} and \ref{fig:HANA_example}).
However, we also find large performance gains for the US census, particularly for the small database, which seems intuitive as smaller datasets have less information to learn from and thus would benefit more from transfer learning. 

\vspace{0.2cm}

The US census images differ from those of the HANA database and the Danish census in that they contain only the surname.
This might contribute to the smaller performance gain we see when applying transfer learning, compared to the Danish census.
Further, the performance both with and without transfer learning is worse on the US census.
We use a large test sample from the US census to validate the performance with more than 60,000 test examples, while for the Danish census data we have approximately 6,000 test examples. 
In general, it seems that the US census data is more difficult to transcribe, making the reduction in error rates from transfer learning even more promising.
At full data coverage on the Danish census, the WACC increases from 77.8\% to 92.2\% for the small training set and 86.1\% to 94.6\% for the large training set.
On the US census, the WACC increases from 72.8\% to 78.7\% for the small training set and from 84.7\% to 86.8\% for the large training set. 

\vspace{0.2cm}

We believe that transfer learning from the HANA database can provide large gains when transcribing handwritten names from other data sources.
These gains are particularly large when transfer learning to a domain that is close to the HANA database and when relatively few labelled images are available. The gains can also be substantial when transfer learning to a domain that is further away but when relatively many labelled images are available.
Using transfer learning with more than 50,000 training sample points, we achieve an error rate reduction of 61.4\% for the Danish census and 14.2\% for the US census data.
This equates to 21,772 corrections of falsely transcribed images when transcribing one million handwritten US names.
For the Danish names, and for the US names when only 10,000 labelled images are available, the increase in transcription accuracy is much larger.
Thus, while transfer learning leads to smaller gains when more labelled images are available, the benefits are still tangible.
Further, we find that most currently available datasets with handwritten names contain fewer than 50,000 labelled handwritten names, and labelling thousands of images is both time-consuming and expensive.
This means that transfer learning from the HANA database not only help improving transcription accuracy, it could also reduce costs as fewer labelled images will be needed.



%% file: tex/4_discussion.tex
\section{Discussion}
\label{Discussion}
Table \ref{tab:CHAR} and Figures \ref{fig:performance0} and  \ref{fig:performance1} summarize our results on the HANA database. Due to computational constraints, we only tested the performance of relatively few models. 
Yet, our models still achieved impressive performance, being able to transcribe names with high accuracy.
As these models are the first results on this database, there are currently no available comparable results, and we hope that other researchers can use these results as a benchmark and transfer learn from this database.
To show the validity of such a strategy, we presented two transfer learning exercises in detail (with a third one discussed in Appendix~\ref{app: Appen}), showing that the use of the HANA database can significantly increase the transcription accuracy of names from both Danish and US census.

\vspace{0.2cm}

We believe that large-scale databases are a necessary prerequisite for achieving high accuracy when transcribing handwritten text. This database proves to be sufficiently large for models to read handwritten names with high accuracy. The high performance is achieved despite several stated complications. The most common complications with the labels and the corresponding images are the structure of the personal names on the images relative to the labels, confusion of certain characters, and general typos. We emphasize that the labels are not perfect and we find that this is especially true for harder to read cases where certain characters are ``open to interpretation''.

\vspace{0.2cm}

As a robustness check one could also test the models using phonetically spelled versions of the names, e.g., \textit{Christian} versus \textit{Kristian}. We choose not to do this in our benchmark models as there exist labels that are very similar but have different meanings. Therefore, by allowing for small discrepancies in the names one could easily create mislabeled training data across very similar names. We realize that it could to some extent mitigate the complication from the harder to read cases where the transcribers possibly made mistakes, but we leave this as an open question for future work.

%% file: tex/5_conclusion.tex
\section{Conclusion}
\label{Conclusion}
This paper introduces the HANA database, which is the largest publicly available handwritten personal name database. The large-scale HANA database is based on Danish police register sheets, which have been made freely available by Copenhagen Archives. The final processed database contains a total of 3,355,033 names distributed across 1,105,904 images. Benchmark results for transcription based deep learning models are provided for the database on the last name, first and last name, and full name. 

\vspace{0.2cm}
Our goal is to create and promote a more challenging database that in many ways is more comparable to other historical documents. Specifically, historical documents are often tabulated and can therefore be cropped into single-line fragments, which should make it easier to train HTR models and to make more efficient transcriptions. Second, the naturalism of the police register sheets are in our opinion quite comparable to a lot of widely used historical documents such as census lists, parish registers, and funeral records. This makes any performance based on these documents more representative of the performance that would be obtained in custom applications. To validate this point, we showed examples of models transfer learning from the HANA database on Danish and US census handwritten names. We find that transfer learning increases the word accuracy from 77.8\% to 92.2\% (86.1\% to 94.6\%) for the Danish census and from 72.8\% to 78.7\% (84.7\% to 86.8\%) for the US census when 10.000 (50.000) training examples are available.

\vspace{0.2cm}
We want to highlight two important features of our database. First, despite the challenges associated with labelling errors and unstructured images, the size of the database appears to compensate, making possible high performance models for automatically transcribing handwritten names. Second, related to the prior point, despite the commonness of names being far from evenly distributed, resulting in a highly unbalanced sample of the represented names, with 65,020 names singularly represented out of a total of 105,607 different names, the models still generalize well. We view this as very encouraging, suggesting that high performance automatic transcription is possible even in difficult and realistic scenarios. 

\vspace{0.2cm}
We have performed image-processing procedures to make the database useful for training single-line learning systems. Further, the code for replicating our results and transfer learning from our models is made freely available. We strongly encourage other researchers to use the HANA database and to make improvements to our procedures in order to continuously increase the size and quality of the database.  Ultimately, we believe this can help making automatic transcriptions of personal names and other handwritten entities much more precise and cost efficient in addition to making the transcriptions fully end-to-end reproducible. By adding improvements to existing linking methods, due to fewer transcription error rates, this could further incentivize the usage and construction of reliable long historical databases across multiple generations. 
\vspace{0.2cm}

%% file: tex/7_Appendix.tex
\begin{appendix}
\section{Additional Transfer Learning Illustration}
\label{app: Appen}
As an additional illustration, we transfer learn from our last name model to the Handwriting Recognition database available from Kaggle containing roughly 410,000 images. To do so, we make a few changes to make the data fit into our current framework. First, we split names including hyphens and only include the last part of the names. This means that 806 labels in the test set are altered in this manner, potentially upward biasing our models. In addition, we remove empty name labels and names including  special characters, which reduces the test set from 41,370 to 41,264 images. 
Finally, we remove three images from the training set and one image from the validation set, due to above 18 characters in the corresponding names.
In total, 329,982 training images, 41,252 validation images, and 41,264 test images remain after these corrections. 

\vspace{0.2cm}

The purpose of showing the performance of transfer learning from the HANA database to this dataset is to show the performance gain from transfer learning in a setting with a very large training sample consisting of hundreds of thousands of sample points.
We create two training sets: A ``small'' set, where only the validation images are used for training (41,252 images) and a ``large'' set, where both the training \textit{and} the validation images are used for training (371,234 images).
In both cases, we use the test images to evaluate our models.
We train two models for each set: One where we transfer learn from the HANA database last name model and one where we do not use transfer learning.

\vspace{0.2cm}

We proceed similarly to the transfer learning examples discussed in Section~\ref{Results}.
The only differences are: (1) the image size is smaller, here around 388 by 40, which is the resolution we train at, and; (2) for computational reasons, we conduct only a search for the learning rate for the two models trained on the small set, and then use the learning rates found here also for the two models trained on the large set. 

\vspace{0.2cm}

As we expect, the performance increase of using transfer learning is highest for the small training set, where accuracy increases from 81.33\% to 83.24\%. 
For the large training set, accuracy increases from 87.48\% to 87.83\%.
While these increases are lower compared to our other transfer learning exercises, it is important to note that the size of the training sets are much larger: The ``small'' training set we use for this example is almost as large as the large training set we use in our other examples, and the ``large'' training set we use for this exercise contains 371,234 samples, a scale often not realistically obtained.

\vspace{0.2cm}

\begin{figure}[H]
  \caption{\\Transfer Learning Performance on Kaggle HTR}
  \label{fig: tl-performance-appen}
\centering
\captionsetup{justification=justified}
  \includegraphics[width=0.5\textwidth]{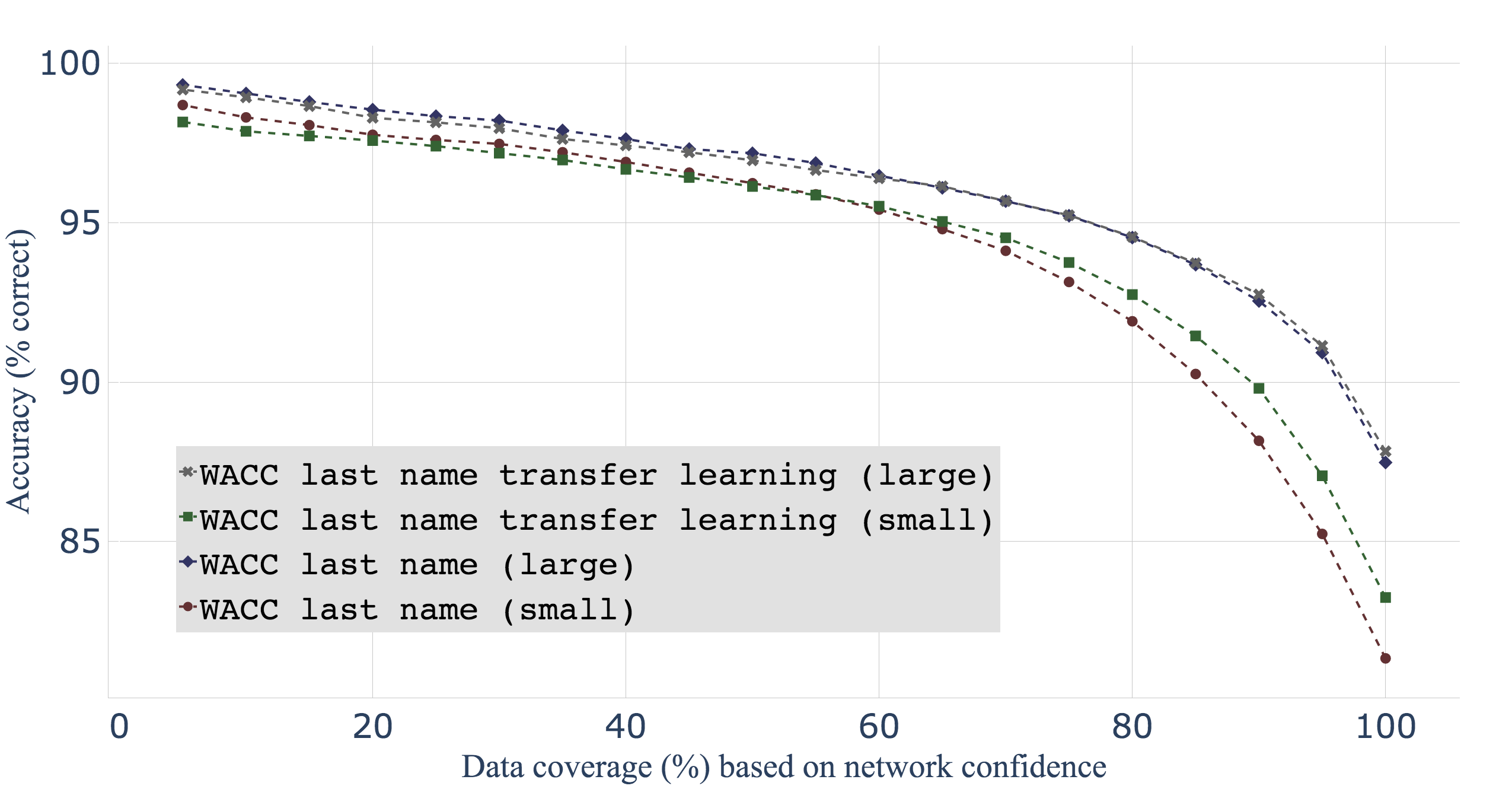}
  \caption*{\scriptsize{The figure shows the performance gain from adopting a transfer learning strategy based on the HANA database.
We find that the performance gain is larger for the smaller training sets, but still present even when nearly 400,000 observations for training are available.
  }}
\end{figure}

We also study the performance of these models at different levels of data coverage.
Figure~\ref{fig: tl-performance-appen} shows the performance of the four models at different levels of data coverage.
Much in line with our earlier results, the models trained on the small set perform worse than the models trained on the large set, but in both cases the model transfer learning from the HANA database outperforms the model not utilizing the HANA database for transfer learning.

\section{Further Characteristics of the Database}
\label{Appendix}

\begin{figure*}
\centering
\caption{\\Further Database Characteristics}
\label{fig:distributions}
\subfloat[Distribution of Names Per Image]{%
  \label{fig:distributions-a}
  \includegraphics[width=0.4\textwidth,valign=t]{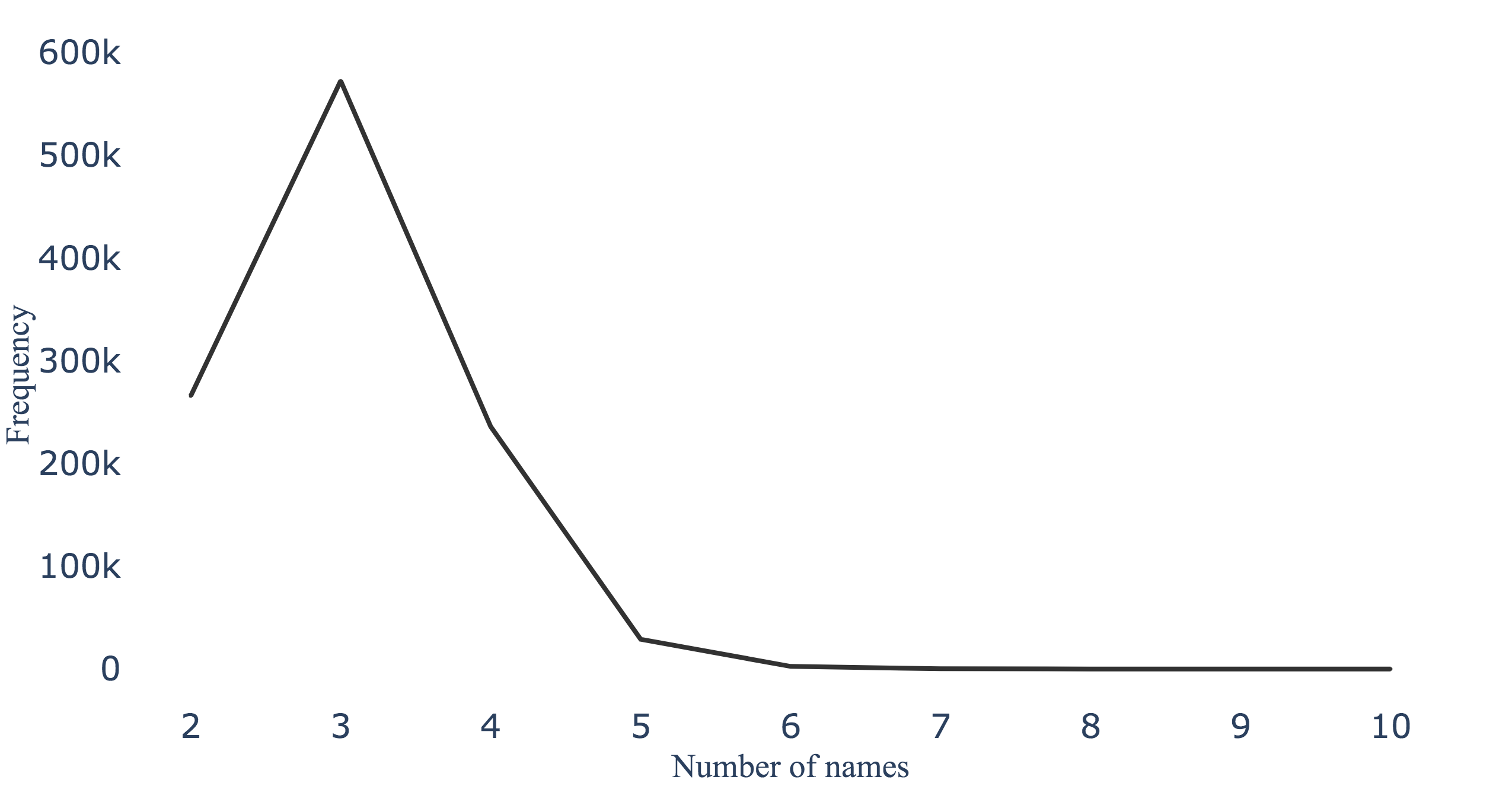}%
}\quad
\subfloat[Distribution of Length of Names]{%
  \label{fig:distributions-b}
  \includegraphics[width=0.4\textwidth,valign=t]{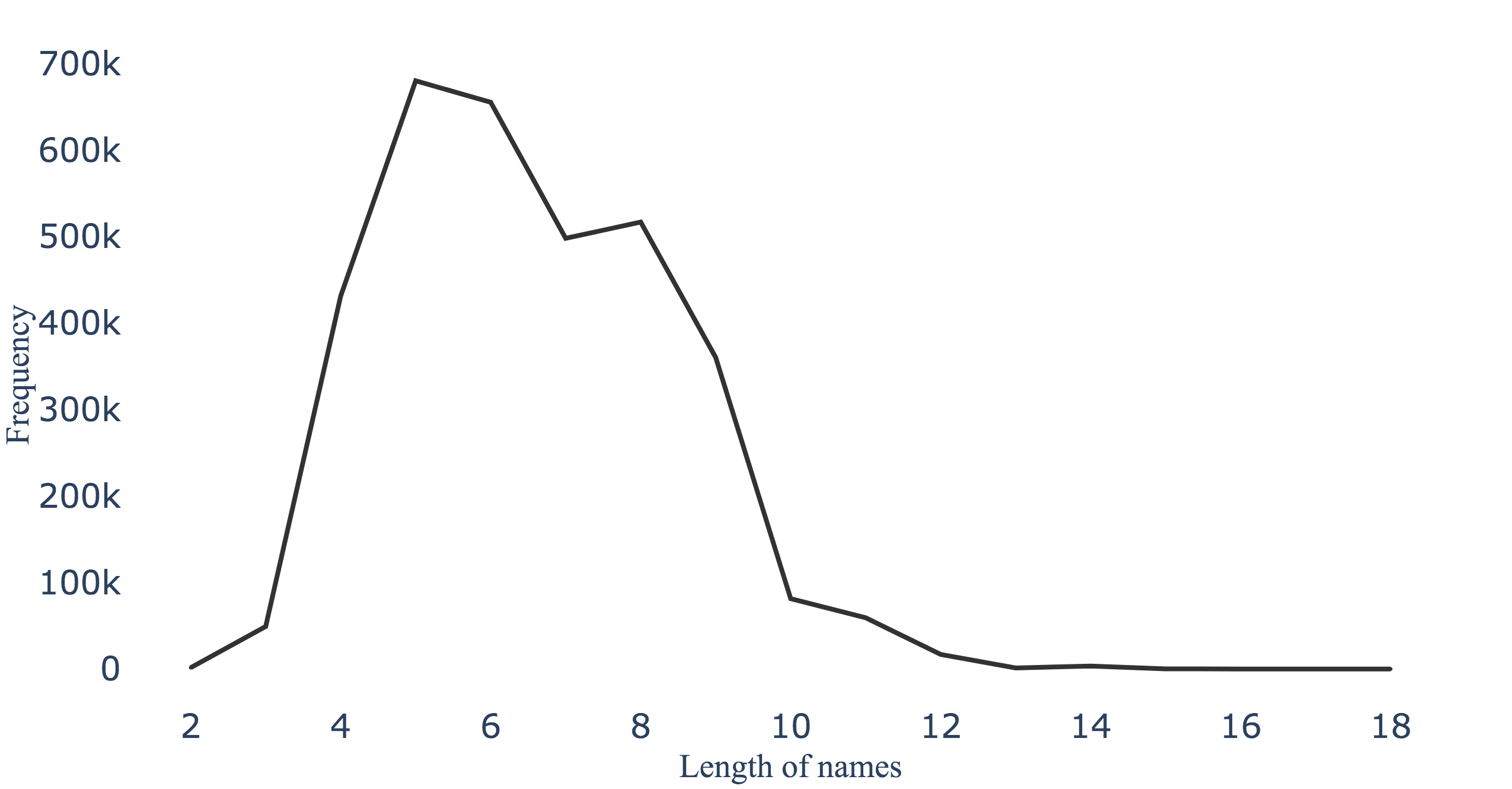}%
}\quad
\subfloat[Distribution of Characters]{%
  \label{fig:distributions-c}
  \includegraphics[width=0.4\textwidth,valign=t]{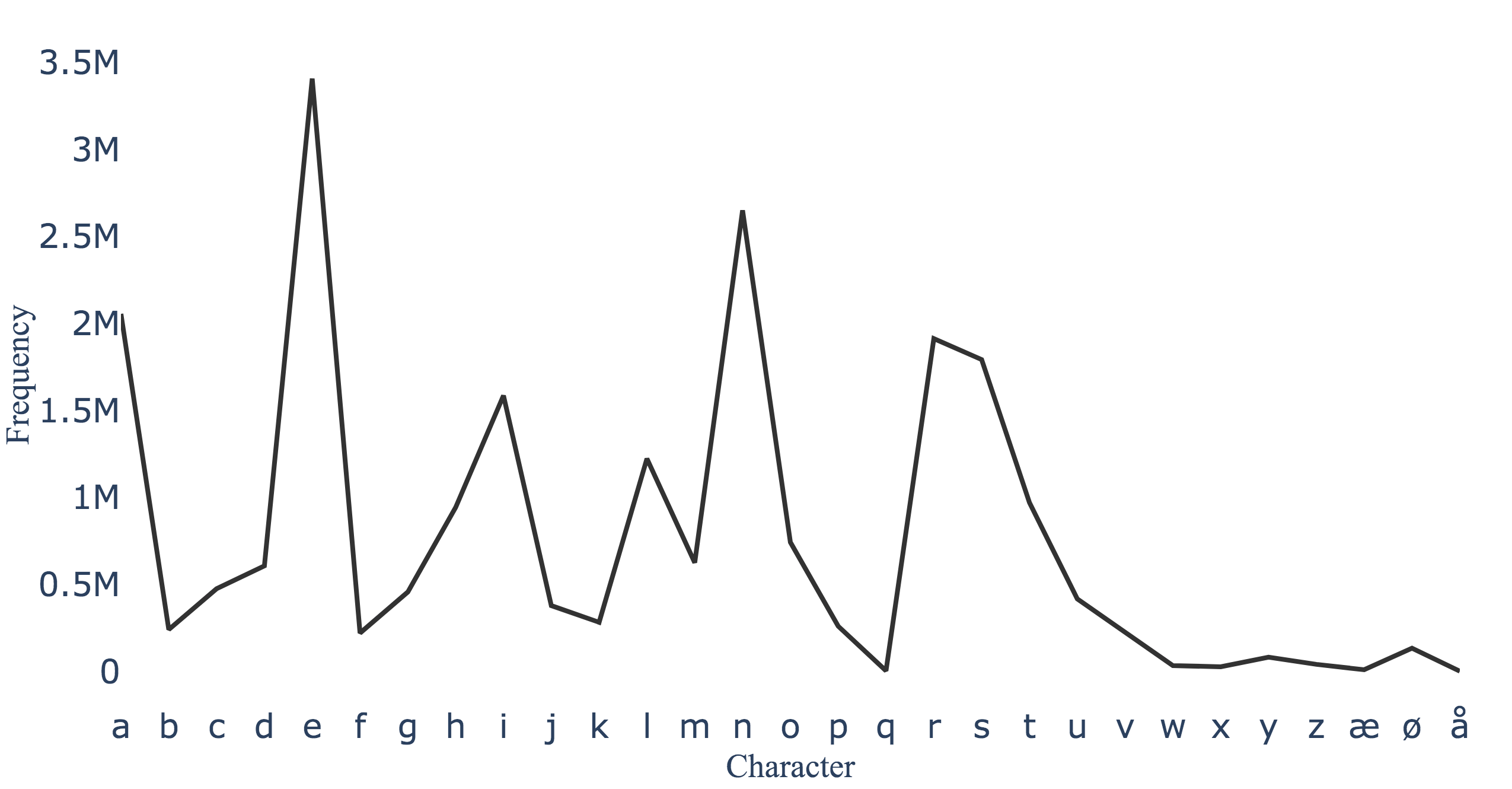}%
}\quad
\captionsetup{justification=justified}
\vspace{-0.2cm}
\caption*{\scriptsize{Panel~\ref{fig:distributions-a} shows the distribution of names per image file. 
As shown, the majority of images contain two to four names; the longest full name consists of 10 separate names.
Panel~\ref{fig:distributions-b} shows the length of names per image. 
The name is in this context defined as each word in a full name, i.e. either first, middle, or last name. 
The longest name consists of 18 characters with most names being somewhere between 3 and 10 characters long. 
Panel~\ref{fig:distributions-c} shows the distribution of characters in the names.
As seen, the most frequent character used in the names is \textit{e}, which appears approximately 3.4 million times, while both \textit{q} and \textit{å} appear fewer than 5,000 times.
All panels aggregates across all individuals present in either the train or test data.
}}
\end{figure*}

In this appendix, we present additional characteristics of the HANA database.
Figure \ref{fig:distributions} shows the distribution of individual names for each full name, the length of the names, and the distribution of characters.
Panel~\ref{fig:distributions-a} shows the number of names per image. 
While most individuals have either just a first and last name, potentially in combination with a single middle name, a significant number have more than one middle name. 
Panel~\ref{fig:distributions-b} shows the length of each name.
This should be interpreted as being the length of either the first, middle, or last name. 
The longest single name in our database is 18 characters. 
It is important to note that this figure does not represent the full name sequence length, as this could potentially be 10x18 characters long. 
The final distribution plot in Panel~\ref{fig:distributions-c} shows the character distribution aggregating across all names.

\end{appendix}

%% file: main_database.bbl
\newcommand{\noop}[1]{}
\begin{thebibliography}{}

\bibitem[\protect\citeauthoryear{Abramitzky, Boustan, and Eriksson}{Abramitzky
  et~al.}{2012}]{Abramitzkyetal2012}
Abramitzky, R., L.~P. Boustan, and K.~Eriksson (2012).
\newblock Europe’s tired, poor, huddled masses: Self-selection and economic
  outcomes in the age of mass migration.
\newblock {\em American Economic Review\/}~{\em 102}, 1832–1856.

\bibitem[\protect\citeauthoryear{Abramitzky, Boustan, and Eriksson}{Abramitzky
  et~al.}{2013}]{Abramitzkyetal2013}
Abramitzky, R., L.~P. Boustan, and K.~Eriksson (2013).
\newblock Have the poor always been less likely to migrate? {E}vidence from
  inheritance practices during the age of mass migration.
\newblock {\em Journal of Development Economics\/}~{\em 102}, 2–14.

\bibitem[\protect\citeauthoryear{Abramitzky, Boustan, and Eriksson}{Abramitzky
  et~al.}{2014}]{Abramitzkyetal2014}
Abramitzky, R., L.~P. Boustan, and K.~Eriksson (2014).
\newblock A nation of immigrants: Assimilation and economic outcomes in the age
  of mass migration.
\newblock {\em Journal of Political Economy\/}~{\em 122}.

\bibitem[\protect\citeauthoryear{Abramitzky, Boustan, and Eriksson}{Abramitzky
  et~al.}{2016}]{Abramitzkyetal2016}
Abramitzky, R., L.~P. Boustan, and K.~Eriksson (2016).
\newblock Cultural assimilation during the age of mass migration.
\newblock {\em NBER Working Paper No. w22381\/}.

\bibitem[\protect\citeauthoryear{Abramitzky, Boustan, Eriksson, Feigenbaum, and
  Pérez}{Abramitzky et~al.}{2020}]{Abramitzkyetal2020a}
Abramitzky, R., L.~P. Boustan, K.~Eriksson, J.~Feigenbaum, and S.~Pérez
  (2020).
\newblock Automated linking of historical data.
\newblock {\em Journal of Economic Literature\/}.

\bibitem[\protect\citeauthoryear{Abramitzky, Mill, and Pérez}{Abramitzky
  et~al.}{2020}]{Abramitzkyetal2020b}
Abramitzky, R., R.~Mill, and S.~Pérez (2020).
\newblock Linking individuals across historical sources: A fully automated
  approach.
\newblock {\em Historical Methods: A Journal of Quantitative and
  Interdisciplinary History\/}~{\em 53}, 94--111.

\bibitem[\protect\citeauthoryear{Bailey, Cole, Henderson, and Massey}{Bailey
  et~al.}{2020}]{Baileyetal2020}
Bailey, M., C.~Cole, M.~Henderson, and C.~Massey (2020).
\newblock How well do automated linking methods perform? {L}essons from {U.S.}
  historical data.
\newblock {\em Journal of Economic Literature\/}~{\em 58}, 997--1044.

\bibitem[\protect\citeauthoryear{Besl and McKay}{Besl and
  McKay}{1992}]{registration}
Besl, P.~J. and N.~D. McKay (1992).
\newblock Method for registration of 3-d shapes.
\newblock {\em Sensor fusion IV: {C}ontrol paradigms and data
  structures\/}~{\em 1611}, 586--606.

\bibitem[\protect\citeauthoryear{{Copenhagen Archives}}{{Copenhagen
  Archives}}{2022a}]{stadsarkiv_mandtal}
{Copenhagen Archives} (2022a).
\newblock Politiets mandtaller.
\newblock
  \url{https://kbharkiv.dk/brug-samlingerne/kilder-paa-nettet/politiets-mandtaller}.
\newblock Accessed: 2022-08-02.

\bibitem[\protect\citeauthoryear{{Copenhagen Archives}}{{Copenhagen
  Archives}}{2022b}]{stadsarkiv}
{Copenhagen Archives} (2022b).
\newblock Politiets registerblade.
\newblock
  \url{https://kbharkiv.dk/brug-samlingerne/kilder-paa-nettet/politiets-registerblade}.
\newblock Accessed: 2022-08-02.

\bibitem[\protect\citeauthoryear{Cubuk, Zoph, Shlens, and Le}{Cubuk
  et~al.}{2020}]{cubuk2020randaugment}
Cubuk, E.~D., B.~Zoph, J.~Shlens, and Q.~V. Le (2020).
\newblock {RandAugment}: {Practical} automated data augmentation with a reduced
  search space.
\newblock In {\em Proceedings of the IEEE/CVF Conference on Computer Vision and
  Pattern Recognition Workshops}, pp.\  702--703.

\bibitem[\protect\citeauthoryear{Dahl, Johansen, Sørensen, Westermann, and
  Wittrock}{Dahl et~al.}{2021}]{dahl2020}
Dahl, C.~M., T.~S.~D. Johansen, E.~N. Sørensen, C.~E. Westermann, and
  S.~Wittrock (2021).
\newblock Applications of machine learning in document digitisation.
\newblock {\em arXiv preprint arXiv:2102.03239\/}.

\bibitem[\protect\citeauthoryear{{Danish National Archives}}{{Danish National
  Archives}}{2022}]{deathregister}
{Danish National Archives} (2022).
\newblock Nyttige hjælpemidler, når du arbejder med kirkebøger.
\newblock
  \url{https://www.sa.dk/da/hjaelp-og-vejledning/nyttige-hjaelpemidler-naar-du-arbejder-%20kirkeboeger/}.
\newblock Accessed: 2022-08-02.

\bibitem[\protect\citeauthoryear{Deng, Dong, Socher, Li, Li, and Fei-Fei}{Deng
  et~al.}{2009}]{deng2009imagenet}
Deng, J., W.~Dong, R.~Socher, L.-J. Li, K.~Li, and L.~Fei-Fei (2009).
\newblock {ImageNet}: {A} large-scale hierarchical image database.
\newblock In {\em 2009 IEEE conference on computer vision and pattern
  recognition}, pp.\  248--255. IEEE.

\bibitem[\protect\citeauthoryear{Feigenbaum}{Feigenbaum}{2018}]{Feigenbaum2018}
Feigenbaum, J.~J. (2018).
\newblock Multiple measures of historical intergenerational mobility: Iowa 1915
  to 1940.
\newblock {\em The Economic Journal\/}~{\em 128\/}(612), F446--F481.

\bibitem[\protect\citeauthoryear{Goodfellow, Bulatov, Ibarz, Arnoud, and
  Shet}{Goodfellow et~al.}{2013}]{goodfellow2013multi}
Goodfellow, I.~J., Y.~Bulatov, J.~Ibarz, S.~Arnoud, and V.~Shet (2013).
\newblock Multi-digit number recognition from street view imagery using deep
  convolutional neural networks.
\newblock {\em arXiv preprint arXiv:1312.6082\/}.

\bibitem[\protect\citeauthoryear{Harris and Stephens}{Harris and
  Stephens}{1988}]{harris}
Harris, C.~G. and M.~Stephens (1988).
\newblock A combined corner and edge detector.
\newblock {\em Alvey vision conference\/}~{\em 15\/}(50), 10--5244.

\bibitem[\protect\citeauthoryear{He, Zhang, Ren, and Sun}{He
  et~al.}{2016}]{he2016deep}
He, K., X.~Zhang, S.~Ren, and J.~Sun (2016).
\newblock Deep residual learning for image recognition.
\newblock In {\em Proceedings of the IEEE conference on computer vision and
  pattern recognition}, pp.\  770--778.

\bibitem[\protect\citeauthoryear{Kim}{Kim}{2020}]{RandAugmentGitHub}
Kim, I. (2020).
\newblock pytorch-randaugment.
\newblock \url{https://github.com/ildoonet/pytorch-randaugment}.
\newblock Accessed: 2022-08-02.

\bibitem[\protect\citeauthoryear{Konow}{Konow}{2009}]{folkeregister}
Konow, L. (2009).
\newblock folkeregister.
\newblock \url{https://denstoredanske.lex.dk/folkeregister }.
\newblock Accessed: 2022-08-02.

\bibitem[\protect\citeauthoryear{Massey}{Massey}{2020}]{Massey2017}
Massey, C.~G. (2020).
\newblock Playing with matches: An assessment of accuracy in linked historical
  data.
\newblock {\em Historical Methods: A Journal of Quantitative and
  Interdisciplinary History\/}~{\em 50}, 129--143.

\bibitem[\protect\citeauthoryear{Myronenko and Song}{Myronenko and
  Song}{2010}]{CPD}
Myronenko, A. and X.~Song (2010).
\newblock Point set registration: {C}oherent {P}oint {D}rift.
\newblock In {\em 2010 IEEE transactions on pattern analysis and machine
  intelligence}, pp.\  2262--2275. IEEE.

\bibitem[\protect\citeauthoryear{Paszke, Gross, Massa, Lerer, Bradbury, Chanan,
  Killeen, Lin, Gimelshein, Antiga, et~al.}{Paszke
  et~al.}{2019}]{paszke2019pytorch}
Paszke, A., S.~Gross, F.~Massa, A.~Lerer, J.~Bradbury, G.~Chanan, T.~Killeen,
  Z.~Lin, N.~Gimelshein, L.~Antiga, et~al. (2019).
\newblock {PyTorch}: {An} imperative style, high-performance deep learning
  library.
\newblock In {\em Advances in neural information processing systems}, pp.\
  8026--8037.

\bibitem[\protect\citeauthoryear{Price, Buckles, Van~Leeuwen, and Riley}{Price
  et~al.}{2019}]{Priceetal2019}
Price, J., K.~Buckles, J.~Van~Leeuwen, and I.~Riley (2019).
\newblock Combining family history and machine learning to link historical
  records.
\newblock {\em NBER Working Paper No. w26227\/}.

\bibitem[\protect\citeauthoryear{Sutskever, Martens, Dahl, and
  Hinton}{Sutskever et~al.}{2013}]{sutskever2013importance}
Sutskever, I., J.~Martens, G.~Dahl, and G.~Hinton (2013).
\newblock On the importance of initialization and momentum in deep learning.
\newblock In {\em International conference on machine learning}, pp.\
  1139--1147.

\bibitem[\protect\citeauthoryear{Szeliski}{Szeliski}{2010}]{extracting_lines}
Szeliski, R. (2010).
\newblock {\em Computer vision: {A}lgorithms and applications}.
\newblock Springer Science \& Business Media.

\bibitem[\protect\citeauthoryear{Van~Rossum and Drake}{Van~Rossum and
  Drake}{2009}]{10.5555/1593511}
Van~Rossum, G. and F.~L. Drake (2009).
\newblock {\em Python 3 Reference Manual}.
\newblock Scotts Valley, CA: CreateSpace.

\end{thebibliography}
